\documentclass[floatfix,rmp,twocolumn,twoside]{revtex4}
\usepackage{graphicx}

\bibpunct{[}{]}{,}{n}{}{,}

\usepackage[english]{babel}
\usepackage{verbatim}
\usepackage{subfigure}
\usepackage{varwidth}
\usepackage{graphicx}
\usepackage{alltt}

\makeatletter
\g@addto@macro\@verbatim\small
\makeatother

\begin{document}

\title{The Object Projection Feature Estimation Problem in \\ Unsupervised Markerless 3D Motion Tracking}
\author{Luis~Quesada\\
  Department of Computer Science and Artificial Intelligence, CITIC, University of Granada, \\
  Granada 18071, Spain \\
  \textit{lquesada@decsai.ugr.es} \\ [3mm]
  Alejandro J. Le\'on\\
  Department of Software Engineering, CITIC, University of Granada, \\
  Granada 18071, Spain \\
  \textit{aleon@ugr.es} 
  }

\begin{abstract}

3D motion tracking is a critical task in many computer vision applications.
Existing 3D motion tracking techniques require either a great amount of knowledge on the target object or specific hardware. These requirements discourage the wide spread of commercial applications based on 3D motion tracking.
3D motion tracking systems that require no knowledge on the target object and run on a single low-budget camera require estimations of the object projection features (namely, area and position).
In this paper, we define the object projection feature estimation problem and we present a novel 3D motion tracking system that needs no knowledge on the target object and that only requires a single low-budget camera, as installed in most computers and smartphones.
Our system estimates, in real time, the three-dimensional position of a non-modeled unmarked object that may be non-rigid, non-convex, partially occluded, self occluded, or motion blurred, given that it is opaque, evenly colored, and enough contrasting with the background in each frame.
Our system is also able to determine the most relevant object to track in the screen.
Our 3D motion tracking system does not impose hard constraints, therefore it allows a market-wide implementation of applications that use 3D motion tracking.

\end{abstract}

\maketitle

\section{Introduction}
\noindent

Optical motion tracking, simply called motion tracking in this paper, means continuously locating a moving object in a video sequence.
2D tracking aims at following the image projection of objects that move within a 3D space.
3D tracking aims at estimating all six degrees of freedom (DOFs) movements of an object relative to the camera: the three position DOFs and the three orientation DOFs \cite{Lepetit2005}.

A 3D motion tracking technique that only estimates the three position DOFs (namely moving up and down, moving left and right, and moving forward and backward) is enough to provide a three-dimensional cursor-like input device driver.

Such an input device could be used as a standard 2D mouse-like pointing device that considers depth changes to cause mouse-like clicks.
It also settles the bases for the development of virtual device drivers (i.e. software implemented device drivers, or not hardware device drivers) that consider three-dimensional position coordinates.

Real-time 3D motion tracking techniques have direct applications in several huge niche market areas \cite{Yilmaz2006}: the surveillance industry, which benefits from motion detection and tracking \cite{Kettnaker1999,Collins2001,Greiffenhagen2001}; the leisure industry, which benefits from novel human-computer interaction techniques \cite{Gallo2011,Shotton2011}; the medical and military industries, which benefit from perceptual interfaces \cite{Bradski2000}, augmented reality \cite{Ferrari2001}, and object detection and tracking \cite{Ali2011,Forman2011,Dong2011}; and the automotive industry, which benefits from driver assistance systems \cite{Handmann1998}.

However, existing 3D motion tracking techniques require either a great amount of knowledge on the target object (i.e. the object to be tracked) or specific hardware to perform the tracking.

Some of these 3D motion tracking techniques require a model of the target object. The generation of that model requires intensive training on a corpus of labelled images in order to induce the object model \cite{Broida1990,Aloimonos1991,Gennery1991,Koller1993,David2004,Cootes2001,Cootes2002,Comaniciu2003,Wu2011}. Corpus-based training is directly out of reach for most casual users and developers.

Other 3D motion tracking techniques require the object to be marked with either passive or active markers \cite{Ali2011,Forman2011}. Casual users may find marker calibration unkempt, time-consuming, and hard to accurately perform. Also, active markers are expensive and may discourage casual users and developers from setting up a personal 3D motion tracking system.

Finally, techniques that require specific hardware, such as twin cameras or Microsoft Xbox360 Kinect devices \cite{Shotton2011} can only be set up after an initial disbursement has been made.
That investment could dissuade casual users and developers from setting up a personal 3D motion tracking system.

On the other hand, if the hardware requirements of a 3D motion tracking system are lowered, a zero deployment cost exploitation is possible.
Particularly, a 3D motion tracking system that only requires a single low-budget camera can be implemented in a wide spectrum of computers and smartphones that already have such a capture device installed.

Several constraints arise as a consequence of using a low-budget camera: monocular vision, low image resolution, high noise levels, JPEG compression artifacts, and a maximum frame rate of 30 frames per second.

On top of that, most low-budget cameras automatically adjust the shutter speed to the environment lighting conditions.
This may lead to sudden changes in the brightness level between consecutive frames and changes in the frame rate, which may drop down to 10 frames per second. 
It should be noted that low frame rates, in turn, may cause motion blur.

In this paper, we present a novel 3D motion tracking system \cite{Quesada2011} that needs no training, calibration, nor knowledge on the target object, and only requires a single low-budget camera.

Our 3D motion tracking system estimates, in real time, the three-dimensional position of unknown unmarked objects that may be non-rigid, non-convex, self occluded, partially occluded, or motion blurred, given they are opaque, evenly colored, and enough contrasting with the background in each frame.
Our 3D motion tracking system is able to determine the most relevant object to track in the screen.

Section \ref{sec:back} covers existing low-budget 3D motion tracking techniques and discusses the drawbacks they present.
Section \ref{sec:objpro} defines the object projection feature estimation problem and compares different approaches for solving it, some of them original in this work.
Section \ref{sec:3dmot} presents our 3D motion tracking system.
Section \ref{sec:exps} exposes the experiments performed to our system.
Finally, Section \ref{sec:concfw} summarizes the obtained conclusions and discloses the future work that derives from our research.

\section{Background} \label{sec:back}
\noindent
In this section, we introduce the existing low-budget 3D motion tracking techniques and we comment on their advantages and disadvantages.

Subsection \ref{sec:mttecmod} describes motion tracking techniques based on model matching. Subsection \ref{sec:mttecimg} explains motion tracking techniques based on image feature analysis. Subsection \ref{sec:mttecdraw} summarizes the drawbacks of the studied techniques.

\subsection{Motion Tracking Techniques Based on Model Matching} \label{sec:mttecmod}
\noindent
Model-based motion tracking techniques match a model of the target object with its projection in the image.
In order to induce the target object model, these systems require intensive training with a huge set of labelled images.

Traditional model-based 3D motion tracking techniques \cite{Broida1990,Aloimonos1991,Gennery1991,Koller1993} match the geometrical features of the target object model with the object projection in the image.
These techniques require the object to be rigid and previously modeled.
The two major drawbacks that model-based 3D motion tracking techniques present are that they cannot match partially or self occluded objects, and that when the object surroundings are cluttered, parts of the surroundings may match fragments of the object model and therefore produce wrong results.

The SoftPOSIT algorithm \cite{David2004} is an extension to the geometrical features matching methods that supports partially and self occluded target objects. However, this algorithm still requires the object to be rigid and previously modeled, and it may fail to track objects when their surroundings are cluttered.

Active Appearance Models (AAMs) \cite{Cootes2001} produce object reconstructions from a target object model and match them with the object projection. AAMs require the target object to be rigid and not occluded.
As this technique cannot model the appearance of an object seen from different angles (e.g. an object that is being rotated in front of the camera), both self-occlusion and rotation of the target object produce tracking errors.

View-Based Active Appearance Models \cite{Cootes2002} are an extension to AAMs that model the appearance of the target object as a set of 2D templates corresponding to views from different angles and, in runtime, they select the most suitable template to perform appearance matching with.
The main drawbacks of this technique are that it cannot track non-rigid objects, that they need an even more intensive training in order to model several different views of the target object, and that they need to implement a model switching policy, which is a very complex problem in its own.

On top of that, AAM-based techniques require a high processing time to match the object projection reconstruction with the actual object projection, therefore these techniques are unable to perform real-time tracking.

\subsection{Motion Tracking Techniques Based on Image Feature Analysis} \label{sec:mttecimg}
\noindent
Motion tracking techniques based on image feature analysis do not require a model of the target object. Instead, they follow a non-modeled object through the video sequence by only considering the video stream.

Kernel-based object tracking techniques \cite{Comaniciu2003} allow tracking non-rigid objects by spatially masking the target object projection with an isotropic kernel and applying optimization to a spatially-smooth similarity function. These techniques require a model of the target object, but they allow its induction from the projection of the object seen in the preceding frame.

Extensions to kernel-based object tracking techniques cope with motion blur \cite{Wu2011}, which might be present in low lighting conditions or when the target object is moving fast.

Orthogonal variant moments features-based motion tracking techniques \cite{Martin2010} determine the rotation transformations, the scale transformations, and the translation transformations performed to an object projection between consecutive frames.
This technique cannot cope with several moving objects, as the orthogonal variant moments features from the different moving objects would interfere.

\subsection{Summary of Existing Techniques} \label{sec:mttecdraw}
\noindent
Existing 3D motion tracking techniques impose strong requirements such as the target object to be rigid, marked, not occluded, or already modeled; the background to be uncluttered and still; or the need for an intensive processing to be performed, which does not allow real-time motion tracking.

Furthermore, existing 3D motion tracking techniques that are not trained to track a specific object do not provide a mechanism to automatically determine the target object. Instead, they need the user to select the target object in the video stream. This makes these techniques unable to perform without supervision.

To the best of our knowledge, no existing motion tracking technique follows the approach we now proceed to describe.

\section{The Object Projection Feature Estimation Problem} \label{sec:objpro}
\noindent
Our unsupervised markerless 3D motion tracking technique requires estimating the centroid and the area of the projection of a target object given an edge image and a point inside the object projection (namely, inner point).
The inner point also has to be updated to increase the probabilities of it being inside the object projection in the next frame.
We call this the object projection feature estimation problem.

In this section, we discuss the advantages and disadvantages of several approaches for solving that problem, some of them original in this work.

\begin{figure}
\centering
\includegraphics[scale=1]{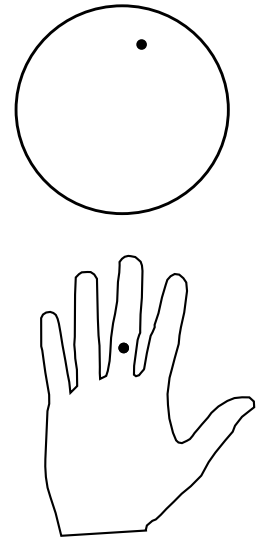}
\caption{The object projection feature estimation problem consists in, given an edge image and a point inside the object projection (namely, inner point), estimating the object projection centroid, the object projection area, and updating the inner point in order to increase the probabilities of it being inside the object projection in the next frame. Example of a convex object projection feature estimation problem (sphere projection) and to a non-convex object projection feature estimation problem (hand projection).}
\label{fig:feat1}
\end{figure}

Figure \ref{fig:feat1} depicts examples of a convex object projection feature estimation problem and a non-convex object projection feature estimation problem.

It should be noted that the inner point can be found enclosed in a small isolated area (e.g. a finger, when the target object is a hand).

It also should be noted that, due to the object movement between frames, it is possible for the current inner point to be relocated at a position that will be outside the object projection in the next frame.

As the inner point determines a position inside the target object projection, and a wrong inner point causes wrong estimations to be performed, the inner point being relocated outside the target object projection is denominated \emph{tracking error}.

Approaches for solving the object projection feature estimation problem cannot determine whether they will cause tracking errors, as no information on the future frames is available when relocating the inner point in the current frame.

Therefore, it is the motion tracking system which has to implement failback strategies that detect the inner point being outside the object projection and relocate it back inside whenever it is possible.

The failback strategies implemented in our 3D motion tracking system are discussed in Section \ref{sec:3dmot}.

Subsections \ref{sec:nray} and \ref{sec:inray} comment on existing aproaches for solving the object projection feature estimation problem.
Subsections \ref{sec:inyray} and \ref{sec:inyrayr} propose extensions to the existing approaches that outperform them.

\subsection{Feature Estimation Based on $n$-Ray-Casting} \label{sec:nray}
\noindent
Using this technique, $n$ rays are casted from the inner point position in different directions to hit an edge in the edge image \cite{Quesada2011}.

The new centroid position is estimated to be the average of the ray hit location positions.

In order to estimate the inner point, it is displaced towards the new centroid until it reaches it or an edge. Then, rays are casted from the inner point and it is relocated at the average of the ray hit location positions, in order to center it in the projection area it is located, which reduces the probability of it being outside the object projection in the next frame.

The area is estimated to be the sum of the lengths of the casted rays.

\begin{figure}
\centering
\includegraphics[scale=1]{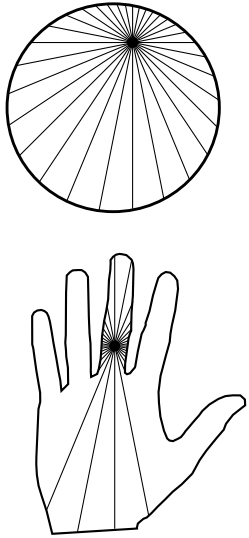}
\caption{$32$-ray-casting being applied to the estimation of the features of a convex object projection (sphere projection) and to a non-convex object projection (hand projection).}
\label{fig:feat2}
\end{figure}

Figure \ref{fig:feat2} illustrates $32$-ray-casting being applied to a convex object projection and to a non-convex object projection.

The main drawback of this technique is that the estimations may not be accurate when it is applied to non-convex object projections (e.g. a hand projection). In that case, the ray hit locations might be representative of just a fragment of the projection, in particular when the inner point is in a small isolated area of the object projection.
The centroid and the area might be inaccurately estimated, and the estimations may greatly vary depending on the position of the inner point relative to the object projection and on the ray orientations.

The likeliness of edge miscalculations (i.e. the edges not being calculated correctly) to have high impact in the projection area and centroid estimations is inversely proportional to $n$.

\subsection{Feature Estimation Based on Iterative $n$-Ray-Casting} \label{sec:inray}
\noindent
Using this technique, $n$ rays are casted from the inner point position in different directions to hit an edge in the edge image \cite{Quesada2011}.

The new centroid position is estimated to be the average of the last iteration ray hit location positions.

The inner point is displaced towards the new centroid until it reaches it or an edge.

The process is repeated until the centroid and inner point adjustment is negligible or up to a maximum number of iterations.

Then, rays are casted from the inner point and it is relocated at the average of the ray hit location positions, in order to center it in the projection area it is located, which reduces the probability of it being outside the object projection in the next frame.

The area is estimated to be the sum of the rays casted during the last iteration.

\begin{figure}
\centering
\includegraphics[scale=1]{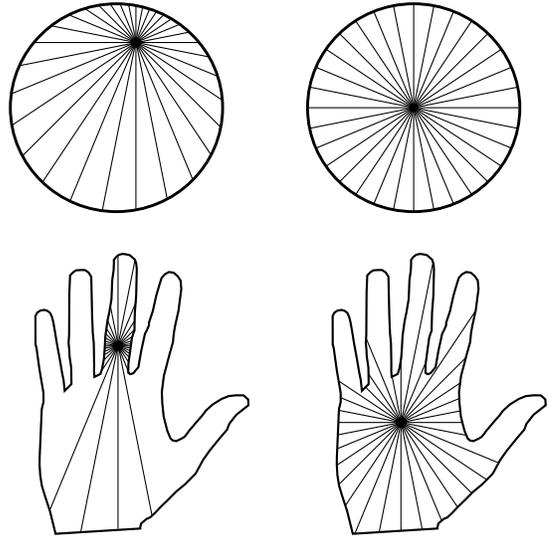}
\caption{Two steps of iterative $32$-ray-casting being applied to the estimation of the features of a convex object projection (sphere projection) and to a non-convex object projection (hand projection). Images on the left show the first iteration. Images on the right show the second iteration.}
\label{fig:feat2it}
\end{figure}

Figure \ref{fig:feat2it} illustrates two steps of iterative $32$-ray-casting being applied to a convex object projection and to a non-convex object projection.

It should be noted that iterative $n$-ray-casting can relocate the inner point into wider areas and therefore produce better estimations of the object projection centroid and area. Indeed, it can be observed that it produces better results than $n$-ray-casting when the target object is non-convex and the inner point is in a small isolated area of the target object projection.

Although this technique being iterative makes the centroid tend to be relocated into wider areas, the estimations are still not accurate when the technique is applied to non-convex object projections, as the ray hit locations might be representative of just a fragment of the object projection.

It should be noted that the centroid is not guaranteed to converge, and the estimations may greatly vary depending on the position of the inner point relative to the object projection, on the ray orientations, and on the maximum number of iterations.

The likeliness of edge miscalculations to have high impact in the projection area and centroid estimations is inversely proportional to $n$.

\subsection{Feature Estimation Based on Iterative $n^y$-Ray-Casting} \label{sec:inyray}
\noindent
We propose iterative $n^y$-ray-casting as an extension to iterative $n$-ray-casting.

Using this technique, $n$ rays are casted from the inner point position in different directions to hit an edge in the edge image.

Then, $n$ rays are casted from each of the last iteration ray hit location position. This re-casting process is repeated $y$ times for a total of $n^y$ rays being casted in the latest iteration.

The new centroid position is estimated to be the average of the last iteration ray hit location positions.

The inner point is displaced towards the new centroid until it reaches it or an edge.

Then, rays are casted from the inner point and it is relocated at the average of the ray hit location positions, in order to center it in the projection area it is located, which reduces the probability of it being outside the object projection in the next frame.

The process is repeated until the centroid and inner point adjustment is negligible or up to a maximum number of iterations.

The area is estimated to be the sum of the rays casted during the last iteration.

\begin{figure}
\centering
\includegraphics[scale=1]{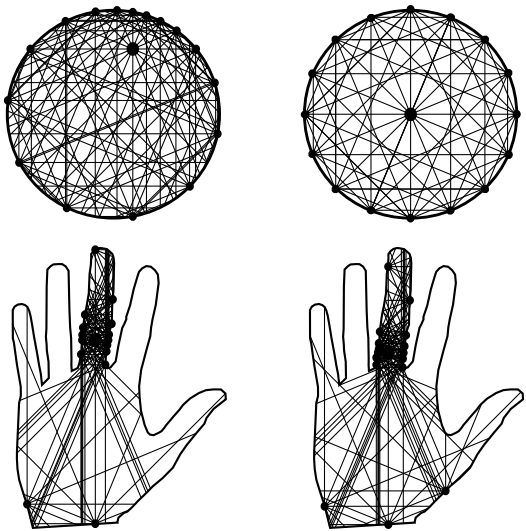}
\caption{Two steps of iterative $16^2$-ray-casting being applied to the estimation of the features of a convex object projection (sphere projection) and to a non-convex object projection (hand projection). Images on the left show the first iteration. Images on the right show the second iteration.}
\label{fig:feat3} 
\end{figure}

Figure \ref{fig:feat3} illustrates $16^2$-ray-casting being applied to a convex object projection and to a non-convex object projection.

It should be noted that the inner point is relocated into wider areas in non-convex object projections very slowly, due to isolated areas near the current inner point having a higher ray-density than wider areas, rendering the later less relevant for the estimation of the projection centroid and area.
On the other hand, iterative $n^y$-ray-casting covers the projection better than iterative $n$-ray-casting, and therefore outperforms it.

It should be noted that this technique, as $n$-ray-casting, does not guarantee the centroid to converge, and results may still greatly vary depending on the position of the inner point relative to the object projection, on the ray orientations, and on the maximum number of iterations.

The likeliness of edge miscalculations to have high impact in the projection area and centroid estimations is inversely proportional $n^y$. It should be noted that edge miscalculations near the inner point may produce very inaccurate results.

\subsection{Feature Estimation Based on Iterative $n^y$-Ray-Casting with $m$-Rasterization} \label{sec:inyrayr}
\noindent
As an extension to iterative $n^y$-ray-casting that solve the aforementioned issues, we also propose iterative $n^y$-ray-casting with $m$-rasterization.

Using this technique, $n$ rays are casted from the inner point position in different directions to hit an edge in the edge image.

Then, $n$ rays are casted from each of the last iteration ray hit location position. This re-casting process is repeated $y$ times for a total of $n^y$ rays being casted in the latest iteration.

Now, a rasterization process takes place. Every $m$x$m$ block that was run through by any of the rays is selected.

The new centroid position is estimated to be the average of the selected block positions.

The inner point is displaced towards the new centroid until it reaches it or an edge.

Then, rays are casted from the inner point and it is relocated at the average of the ray hit location positions, in order to center it in the projection area it is located, which reduces the probability of it being outside the object projection in the next frame.

The process is repeated until the centroid and inner point adjustment is negligible or up to a maximum number of iterations.
It should be noted that, as blocks always represent areas inside the object projection, no blocks are unselected between iterations.

The area is estimated to be the sum of the selected block areas.

\begin{figure}
\centering
\includegraphics[scale=1]{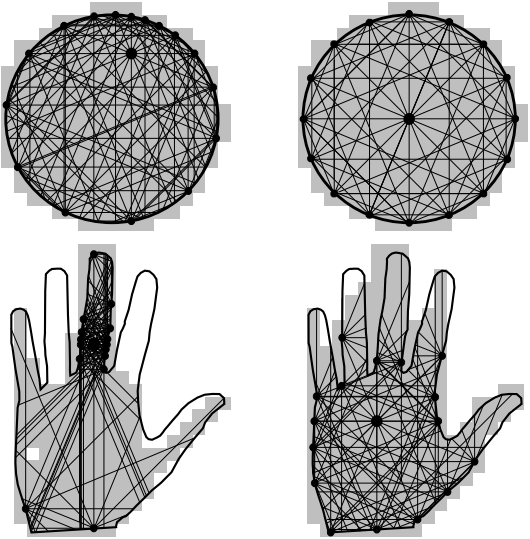}
\caption{Two steps of iterative $16^2$-ray-casting with $8$-rasterization being applied to the estimation of the features of a convex object projection (sphere projection) and to a non-convex object projection (hand projection). Images on the left show the first iteration. Images on the right show the second iteration.}
\label{fig:feat5}
\end{figure}

Figure \ref{fig:feat5} illustrates $16^2$-ray-casting with $8$-rasterization being applied to a convex object projection and to a non-convex object projection.

It should be noted that the inner point moves to wider areas in non-convex object projections quicker than when applying iterative $n^y$-ray-casting, due to high-ray-density areas being given the same relevance as low-ray-density areas. Less iterations are necessary for the estimations to be accurate, therefore processing times are lower than those of iterative $n^y$-ray-casting without rasterization.
It also should be noted that when $m$ is too high, the projection centroid and area estimations will be imprecise due to low resolution in block selection; when $m$ is too low, the technique behaves as iterative $16^2$-ray-casting without rasterization, which makes the inner point to be slowly displaced .

As the selected blocks are kept between iterations, the inner point and the centroid are guaranteed to converge. Although results may vary depending on the position of the inner point relative to the object projection, on the ray orientations, and on the maximum number of iterations, they will be similar for convex object projections and non-convex object projections with not too large isolated areas.

The likeliness of edge miscalculations to have high impact in the projection area and centroid estimations is inversely proportional to $n^y\cdot i$, being $i$ the number of performed iterations, as the final estimations depends on rays casted during any iteration.

This technique produces better results than any of the other studied or proposed techniques.
Therefore, our motion tracking system uses iterative $n^y$-ray-casting with $m$-rasterization to solve the object projection feature estimation problem.

\section{Unsupervised Markerless 3D Motion Tracking} \label{sec:3dmot}
\noindent
In this section, we introduce our proposal, a 3D motion tracking system that only imposes the constraints of the target object being opaque, evenly colored, and enough contrasting with the background in each frame, and that only requires the use of a single low-budget camera.

Our approach does not require the use of markers in the target object, nor a model of target object.

Our 3D motion tracking system accepts as input a stream of frames from a camera and produces as output the three-dimensional coordinates of the target object relative to the camera.

Subsection \ref{sec:fprec} describes how captured frames are preprocessed.
Subsection \ref{sec:sov} provides an overview on the system and implementation details.
Subsection \ref{sec:output} comments on the estimation of the output three-dimensional coordinates.

\subsection{Frame Preprocessing} \label{sec:fprec}

\begin{figure*}[tb]
\centering
\includegraphics[scale=1]{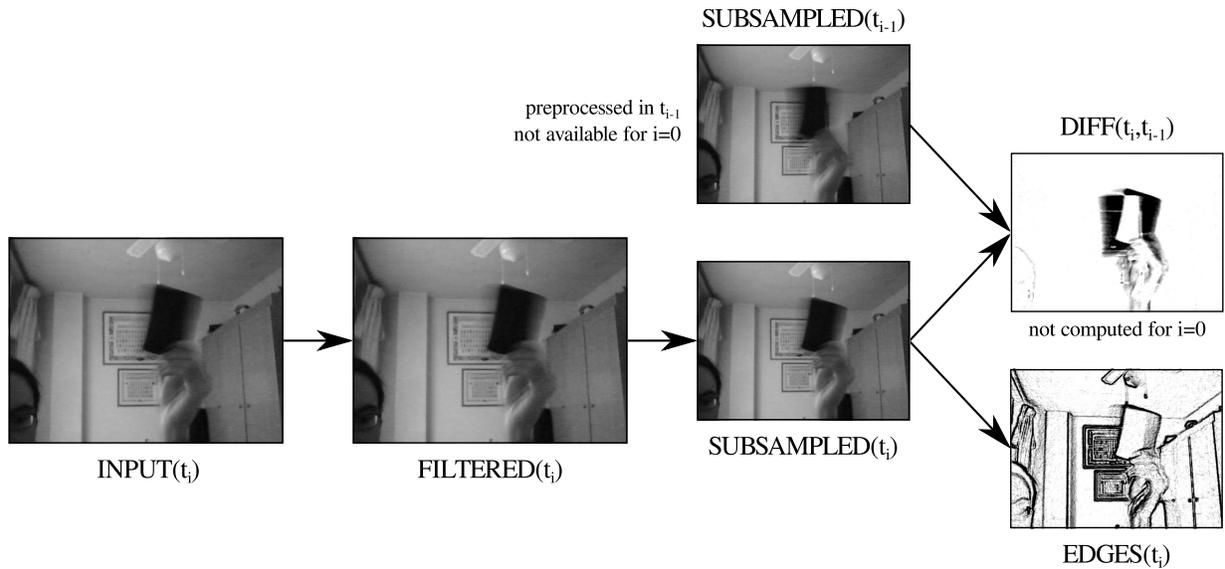}
\caption{Frame preprocessing.}
\label{fig:imagetrans}
\end{figure*}

Each time a frame is captured by the camera, a preprocess that takes as input this current frame, $t_i$, and the previously processed frame, $t_{i-1}$, is performed.
It should be noted that there may not exist previous frame (i.e. $i=0$, which means the current frame is the first ever captured). In that case, the part of the preprocessing that uses the previous frame is omitted.
This preprocessing step is illustrated in Figure \ref{fig:imagetrans}.

The frame is convoluted using a Poisson disk filter of radius 5 that reduces the effect of both JPEG 8x8 block compression artifacts and camera noise.

The filtered frame is subsampled down to 160x120 in order to reduce further processing time and mitigate the effects of motion blur.

An edge detection filter based on the Sobel operator \cite{Pratt2007,Kanopoulos1988} is applied to the frame in order to produce a grayscale edge image.
The edge image will be used during the estimation of the features of the target object projection. Any pixel value over 128 in the edge image is considered an edge.

Using a lower edge threshold would make the system perform correctly when the target object and its surroundings are similarly colored. However, in exchange, it would require the target object to be colored more evenly, which is not desired.
On the other hand, using a higher edge threshold would require the target object to be colored less evenly. Although, in exchange, edge miscalculations would happen when the target object and its surroundings are colored too similarly, which is not desired.

Finally, if a previous frame is available (that is, the current frame is not the first ever captured frame), the current and the previous subsampled frames are compared in order to calculate a absolute difference image.
The absolute difference image will be used to estimate the global amount of movement between frames, which is used to determine if there is a target object in the scene and, in that case, a point inside its projection (i.e. its inner point).

The automatic shutter speed adjustment that low-budget cameras perform can cause sudden changes in the overall brightness level of captured frames. 
These brightness changes affect the captured frames by increasing or decreasing each pixel with a value whose probabilistic distribution depends on the subjacent camera hardware.

That results in the miscalculation of the absolute difference image. A wrong absolute difference image may result in the incorrect determination of the target object.

In order to mitigate this problem, the 10 different minimum pixel values in the absolute difference image are found. The maximum brightness level change is estimated to be the maximum of those pixel values. Each pixel in the square difference image is adjusted by substracting it the maximum brightness level change and flooring it to 0.

This effectively mitigates most of the effects of a sudden brightness change in the absolute difference image.

Experiments have proven that using less than 10 different minimum pixel values may not sufficiently attenuate the overall brightness change in some cases, as the brightness level may affect some pixels more than others. On the other hand, using more than 10 different minimum pixel values seems to remove too much information from the absolute difference image when there was no sudden brightness level change.

The global inter-frame movement value is estimated to be the average of the values of all the pixels in the absolute difference image.

\subsection{System Overview} \label{sec:sov}

\begin{figure}[tb*]
\centering
\includegraphics[scale=1]{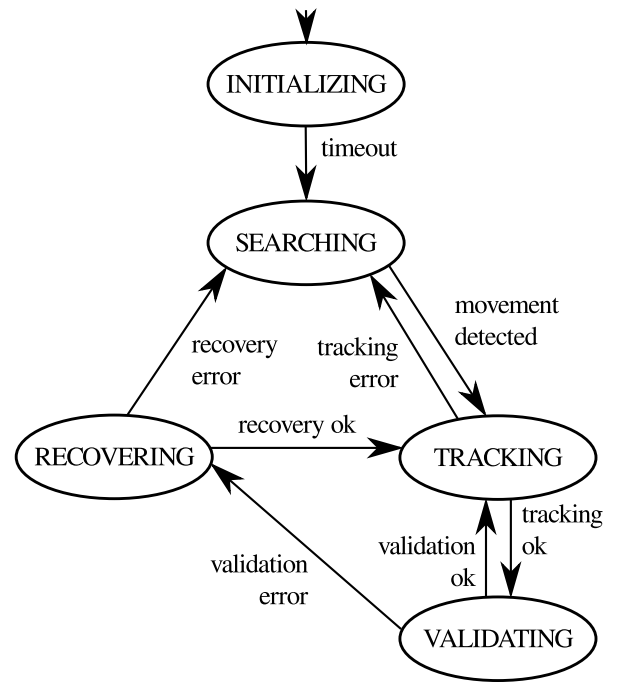}
\caption{The motion tracking system state machine.}
\label{fig:fsm}
\end{figure}

Figure \ref{fig:fsm} summarizes the system behavior in the form of a state machine. We now proceed to explain the behavior of the system when it is in all the different states.

When the system is in \emph{\textbf{INITIALIZING}} state, it is performing a one-time startup process.
The system waits 2 seconds while the camera shutter adjusts its speed, avoiding the flashes that may occur during the automatic shutter speed startup setup that most low-budget cameras perform.
After the 2 seconds have passed, the first frame is read and preprocessed as explained in Subsection \ref{sec:fprec} and the system switches to \emph{SEARCHING} state.

When the system is in \emph{\textbf{SEARCHING}} state, it is waiting for an object to track.
A new frame is read from the input stream and preprocessed as explained in Subsection \ref{sec:fprec}.
If the global inter-frame movement value is higher than $57,600$ (i.e. an average of 3 per pixel), it is determined that a new object has to be tracked.
The centroid of the target object projection and an inner point are estimated to be the center of mass (i.e. the average position weighted by the pixel magnitude) of the absolute difference image, and the system is switched to \emph{TRACKING} state.

When the system is in \emph{\textbf{TRACKING}} state, it is tracking a target object.
A new frame is read from the input stream and preprocessed as explained in Subsection \ref{sec:fprec}.
The target object projection centroid, inner point and area are estimated using $16^2$-ray-casting with $8$-rasterization, as explained in Subsection \ref{sec:inyrayr}. 
The system switches to \emph{VALIDATING} state if the tracking succeeds or to \emph{SEARCHING} state if it fails. The tracking fails in any of the following cases: if the global inter-frame movement value has been below $4,800$ (i.e. an average of 0.25 per pixel) for 2 seconds, which means there is no movement in the scene; or if the object projection area is higher than $60\%$ or lesser than $2\%$ of the screen, which would make it difficult for the tracking algorithm to perform correctly. The tracking succeeds in any other case.

When the system is in the \emph{\textbf{VALIDATING}} state, it is checking if the new target object projection inner point is consistent with the previous frame object projection. Let $CP$ be the average RGB color tuple of the 5x5 square centered in the previous inner point in the previous frame and $CC$ be the average RGB color tuple of the 5x5 square centered in the current inner point in the current frame. If $abs(CP_R-CC_R)+abs(CP_G-CC_G)+abs(CP_B-CC_B)$ is lower or equal to 90 (namely, the previous inner point in the previous frame color-matches the current inner point in the current frame), the system switches to \emph{TRACKING} state.
In any other case, the inner point might have been incorrectly relocated, so the system switches to \emph{RECOVERING} state.

\begin{figure}[b*]
\centering
\includegraphics[scale=1]{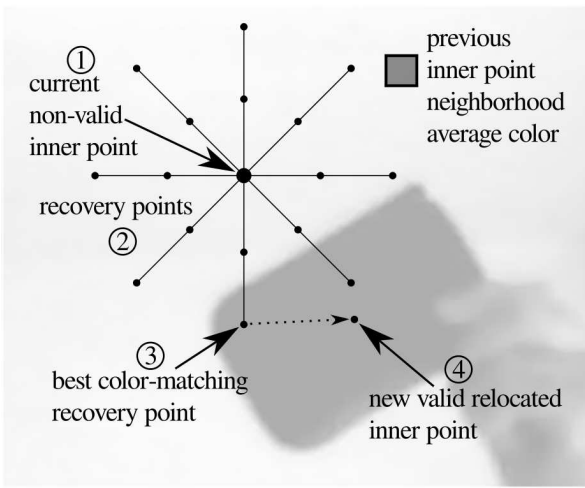}
\caption{Color-matching-based inner point relocation failback strategy.}
\label{fig:colormatch}
\end{figure}

When the system is in \emph{\textbf{RECOVERING}} state, it is trying to relocate the inner point, which may have been incorrectly relocated outside of the target object boundaries.
Figure \ref{fig:colormatch} illustrates the color-matching-based inner point relocation failback strategy, which allows the inner point to be correctly relocated after tracking errors caused by object projection feature estimation errors, fast moving objects, or low frame rates.
A set of recovery points is generated by adding 10 pixel and 20 pixel horizontal, diagonal, and vertical offsets to the current inner point position.
The recovery point that best color-matches the previous inner point in the previous frame, if any, is denominated candidate recovery point.
If there is a candidate recovery point, the target object projection centroid, the inner point and area are estimated using $16^2$-ray-casting with $8$-rasterization, as explained in Subsection \ref{sec:inyrayr}.
If the relocated inner point in the current frame color-matches the previous inner point in the previous frame, the recovery has succeeded and the system switches to \emph{TRACKING} state.
In any other case, the recovery has failed and the system switches to \emph{SEARCHING} state.

Whenever the system is tracking an object, the three-dimensional coordinates can be calculated as explained in the next Subsection.

\subsection{Output Three-Dimensional Coordinate Estimation} \label{sec:output}

The system needs to provide estimations of the $(x,y,z)$ object three-dimensional coordinates from the estimated object projection centroid and area.

The $x$ and $y$ coordinates are estimated to be the coordinates of the object projection centroid.

The $z$ coordinate is estimated to be the square root of the object projection area.

As the actual target object size is unknown, the projection area when the target object starts being tracked is assumed to be the reference 0 depth.
Whenever the tracked target object gets closer to the camera, the depth increases, and whenever the target object gets apart from the camera, the depth decreases.

The output 3D coordinates are smoothed to filter out the effect of possible tracking errors and edge miscalculations. A factor is applied to the 3D coordinates, so, each cycle, their value becomes $90\%$ of their old value and $10\%$ of the new estimation. These values provide a good balance between sensitivity and error absorption capabilities.

The next section exposes the experiments performed on our 3D motion tracking system and discloses the obtained results.

\section{Experimental Results} \label{sec:exps}

We have performed extensive experiments on our 3D motion tracking system in order to test its behavior under several conditions.

All the experiments were run on three video streams of the same scene captured by three different devices: a standard low-budget external USB webcam, a low-budget laptop integrated webcam, and a full HD digital camcorder.
The frame rate of both the low-budget camera video streams was 30 in high lighting conditions and 10 in low lighting conditions. The frame rate of the full HD camera video stream was always 60.

Subsections 5.1 to 5.5 describe the set up experiment and the obtained results.
Subsection 5.6 presents a summary of the results.

\subsection{Ideal Scenario Test}
\noindent
This experiment represents the best-case scenario for our 3D motion tracking system.

The cameras were set up in front of a clean white board.
The illumination was proper (i.e. high lighting conditions).
The target object was a black mate sphere hung with transparent nylon thread.
The thread was moved in order to displace the target object around the scene.

As the light conditions are proper, the camera frame rates are 30 for both the low-budget cameras and 60 for the full HD camera.
The high frame rates caused almost no motion blur, therefore the edges of the target object were clear.
The white background did not interfere with the silhouette of the black target object projection.

The evenly-colored, rigid, convex, and clearly distinguishable target object was correctly detected and tracked in every frame using the three video devices.

\subsection{Low Light Conditions Test}
\noindent
This experiment is designed to determine the effect of low lighting conditions in our 3D motion tracking system.

The cameras were set up in front of a clean white board.
The illumination was improper (i.e. low lighting conditions).
The target object was a black mate sphere hung with transparent nylon thread.
The thread was moved in order to displace the target object around the scene.

As there were low light conditions, low-budget cameras adjusted the shutter speed to 10 frames per second, causing motion blur to appear. The full HD camera frame rate was still 60, although the image was noisier.
Our system supported high image noise levels and compensated motion blur with the failback strategy, and the evenly-colored, rigid, convex, and clearly distinguishable target object was still correctly detected and tracked in most frames using the three video capture devices. The inner point failback strategy had to be applied in some frames of the low-budget camera video streams, in particular when the object was moving fast. The failback strategy always succeeded and it was able to recover the tracking in all cases.

\subsection{Interfering Background Test}
\noindent
This experiment is designed to determine the effect of a cluttered or moving background in our 3D motion tracking system.

The cameras were set up in front of a room with a variety of furniture.
The illumination was proper (i.e. high lighting conditions).
The target object was a black mate sphere held and partially occluded by a hand.
It should be noted that the hand and the arm were also moving in the scene.

The target object was correctly determined to be the black sphere in all the tests, as its contrast with the surroundings caused the center of mass of the difference image to be inside its projection.
The tracking was performed correctly, although edge miscalculations due to the cluttered background caused slight variations in the projection centroid and area estimations.
These variations were negligible in all cases, since they lasted a couple frames at most and the system 3D coordinate smoothing managed to absorb them. The failback strategy did not need to be applied.

Although the full HD camera provided better results mainly due to its higher constant frame rate, the low-budget cameras provided enough accurate results and no there was no loss of tracking.

\subsection{Non-Convex Target Object Projection Test}
\noindent
This experiment is designed to determine the effect of non-convex target object projections in our 3D motion tracking system.

The cameras were set up in front of a clean white board.
The illumination was proper (i.e. high lighting conditions).
The target object is a paper shaped like a hand with the fingers spread, hung with two nylon threads.
Both threads are moved in order to displace the target object around the scene while making it face the cameras.

The paper hand silhouette was correctly detected and tracked in every frame using the three video devices.
Slight orientation changes did not greatly influence the area estimations.
The gaps between the fingers being too close to the inner point caused some tracking errors.
However, the inner point failback strategy managed to fix this situation and return the tracking to the hand in all cases.
The projection area was accurately estimated.

\subsection{Real World Conditions Test}
\noindent
This experiment is designed to determine the accuracy and performance of the system under real world conditions.

The cameras were set up in front of a room with a variety of furniture.
The illumination was proper (i.e. high lighting conditions).
The target object was an actual hand with the fingers spread. The color of the hand reasonably contrasts with the background and with the shirt that covers the arm.

The hand was correctly detected and tracked in most frames using the three video devices.
Tracking problems arose when the hand was moved through similarly colored background areas, in which cases the failback strategy succeeded.

\subsection{Experimental Result Summary}
\noindent
The conclusions we reached after experimentally testing our 3D motion tracking systems are:

\begin{itemize}
\item Objects with convex or non-convex projections are correctly detected and tracked.
\item Objects whose projections are or become self or partially occluded are correctly tracked. The estimation of their projection area decreased proportionally to the occluded zone.
\item The failback strategy is able to effectively detect tracking errors and relocate the inner point in most cases.
\item When the target object is moving through a cluttered background area with which it does not contrasts enough, the centroid location and the area estimation can be slightly off. However, the system 3D coordinate smoothing robustly absorbs the offset.
\item Precise results are obtained for frame rates of 10, 30, and 60. The higher the frame rate, the less motion blur is captured, therefore the sharper the edges look and, consequently, the more accurate the motion tracking is.
\item Noise in the video stream does not significantly affect the results.
\end{itemize}

Our 3D motion tracking system proved to be robust and flexible enough for real world applications.

In the next section, we summarize our work and disclose the future work that derives from our research.

\section{Conclusions and Future Work} \label{sec:concfw}

Existing 3D motion tracking techniques require either a great amount of knowledge on the object to be tracked, or specific hardware to perform the tracking. These requirements discourage the wide spread of applications that use 3D motion tracking.

We have defined the object projection feature estimation problem, which is ubiquitous in unsupervised markerless 3D motion tracking systems.

We have studied existing approaches for solving the object projection feature estimation problem, and we have proposed extensions to these approaches that outperform them.

We have presented a novel 3D motion tracking system that is able to determine the most relevant object in the screen and estimate its three-dimensional position given it is opaque, evenly colored, and enough contrasting with the background in each frame.

Our system performs 3D motion tracking in real time by analyzing the video stream from a single low-end camera.

Our 3D motion tracking system requires no training, no calibration, no previous knowledge on the target object, and no use of markers in the target object.

The experiments performed on our system proved that it is accurate and robust enough to perform correctly in real world conditions, such as cluttered or moving background; not proper lighting; and target objects being non-rigid, non-convex, partially or self occluded, and motion blurred.

Therefore, our 3D motion tracking system settles the bases for the market-wide implementation of applications that use 3D motion tracking.

We plan to improve the approaches for solving the object feature estimation problem in order to make them faster, more robust and accurate.

We plan to optimize the system so that it can run in low processing power devices such as smartphones.

We also plan to extend our 3D motion tracking technique in order to make it able to track multiple objects in the screen.

Finally, we plan to develop virtual devices that implement new input paradigms. These virtual devices will be applied to the interaction with virtual environments and intelligent virtual environments.

\bibliographystyle{plain}
\bibliography{doc}

\end{document}